%% file: main.tex
\documentclass[letterpaper, 10 pt, conference]{ieeeconf}

\usepackage{graphicx}
\usepackage{amsmath}
\usepackage{amssymb}
\usepackage{booktabs}
\usepackage{multirow}
\usepackage{epstopdf}
\usepackage{capt-of}  %

\usepackage[dvipsnames,table,xcdraw]{xcolor}
\newcommand{\bc}{\mathbf{c}}
\usepackage[ruled,vlined,linesnumbered, boxed]{algorithm2e}

\usepackage{yhmath}
\usepackage{cite}

\makeatletter
\let\NAT@parse\undefined
\makeatother
\usepackage[colorlinks, linkcolor=red, anchorcolor=blue, citecolor=blue]{hyperref}

\usepackage{bm}

\usepackage{tikz}
\usetikzlibrary{arrows.meta, positioning, calc, fit, backgrounds}

\newcommand{\bx}{\boldsymbol{x}}
\newcommand{\bU}{\boldsymbol{U}}
\newcommand{\bu}{\boldsymbol{u}}

\newcommand{\bI}{\boldsymbol{I}}

\newcommand{\bg}{\boldsymbol{g}}
\newcommand{\bz}{\boldsymbol{z}}

\providecommand{\coloneqq}{\mathrel{:=}}
\usepackage{mathtools}

\IEEEoverridecommandlockouts

\title{\LARGE \bf
Multi-Modal Trajectory Optimization via \\ Training-Free Reverse Diffusion
}

\author{Tianyu Liu, Wen Yang, Haoran Ding, Yoshihiko Nakamura, and Anqing Duan%
\thanks{All authors are with Mohamed bin Zayed University of Artificial Intelligence (MBZUAI), Abu Dhabi, UAE.}%
}

\begin{document}

\makeatletter
\let\ieee@oldmaketitle\@maketitle
\renewcommand{\@maketitle}{%
  \ieee@oldmaketitle
  \input{sections/teaser_snapshots}}
\makeatother

\maketitle
\thispagestyle{empty}
\pagestyle{empty}

\input{sections/abstract}

\input{sections/introduction}

\input{sections/related_work}

\input{sections/methodology}

\section{Experiments}
\label{sec:experiments}

\input{sections/exp_gmm}

\input{sections/exp_pendulum}

\input{sections/exp_humanoid}

\section{Conclusion and Future Work}
\label{sec:conclusion}

We showed that training-free reverse diffusion, driven solely by
physics-simulator rollouts, can serve as a principled global explorer
for multi-modal trajectory optimization---bridging the gap between
the local convergence of classical planners and the data dependence
of learned methods. Several directions remain open: improving
\emph{sampling efficiency} through variance-reduction techniques or
learned proposal distributions would reduce the number of rollouts
required per diffusion step; exploring \emph{adaptive noise
schedules} that respond to the local cost landscape could accelerate
convergence and further improve mode coverage; and transferring the
pipeline to \emph{physical robotic hardware} will be essential for
validating its robustness under real-world sensing noise, actuator
delays, and model mismatch.

\appendices
\section{Score Estimation under Actuator Box Constraints}
\label{app:box_score}
\input{sections/appendix_box_score}

\bibliographystyle{IEEEtran}
\bibliography{references}

\end{document}

%% file: sections/teaser_snapshots.tex
\vskip 0.8em

\ifdefined\teaserh\else\newlength{\teaserh}\fi
\setlength{\teaserh}{3.8cm}

\begin{center}
  \begin{minipage}[c]{0.63\textwidth}%
    \centering
    \includegraphics[height=\teaserh]{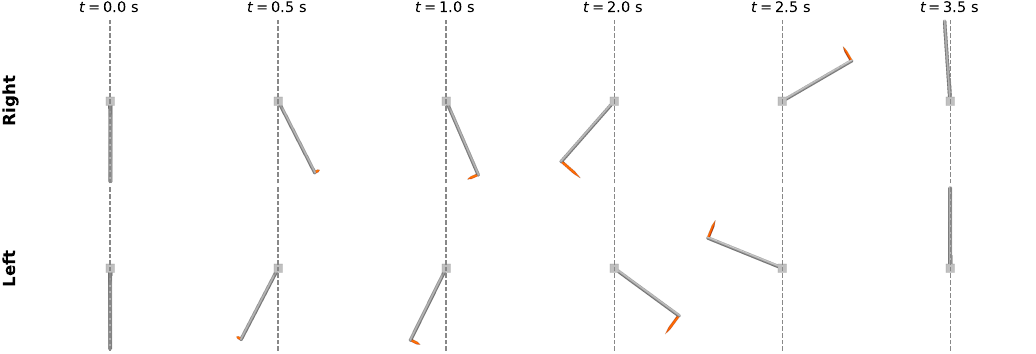}\\[1pt]
    {\footnotesize (a) Pendulum swing-up}
  \end{minipage}%
  \newcommand{\humimg}[2]{%
    \begin{tikzpicture}
      \node[anchor=south west, inner sep=0] (img) {%
        \includegraphics[height=\dimexpr0.5\teaserh-3pt\relax]{#1}};
      \node[anchor=south east, font=\sffamily\bfseries\scriptsize,
            text=green!80!black, scale=0.7]
        at (img.south east) {#2};
    \end{tikzpicture}%
  }%
  \begin{minipage}[c]{0.37\textwidth}%
    \centering
    \begin{minipage}[t]{0.48\textwidth}\centering
      \humimg{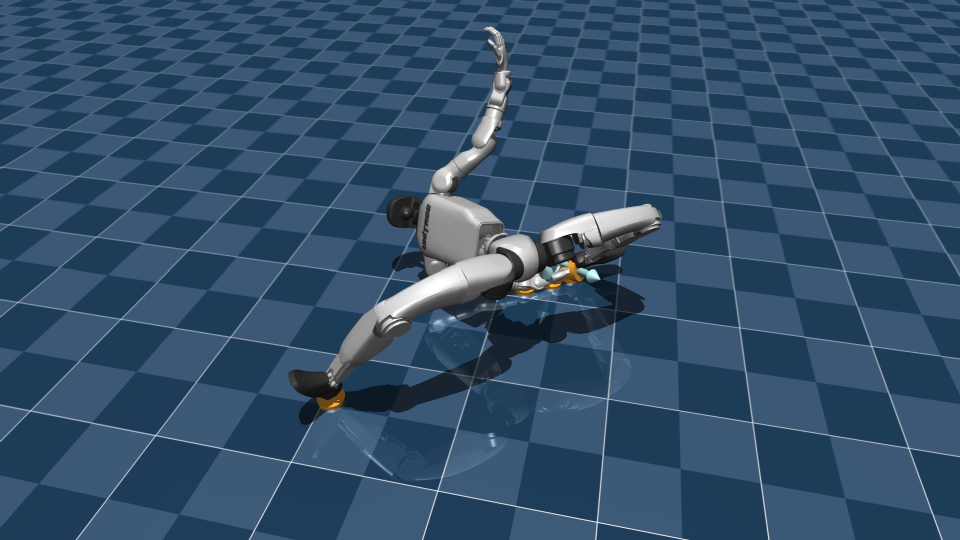}{Hand-Push}
    \end{minipage}\hfill
    \begin{minipage}[t]{0.48\textwidth}\centering
      \humimg{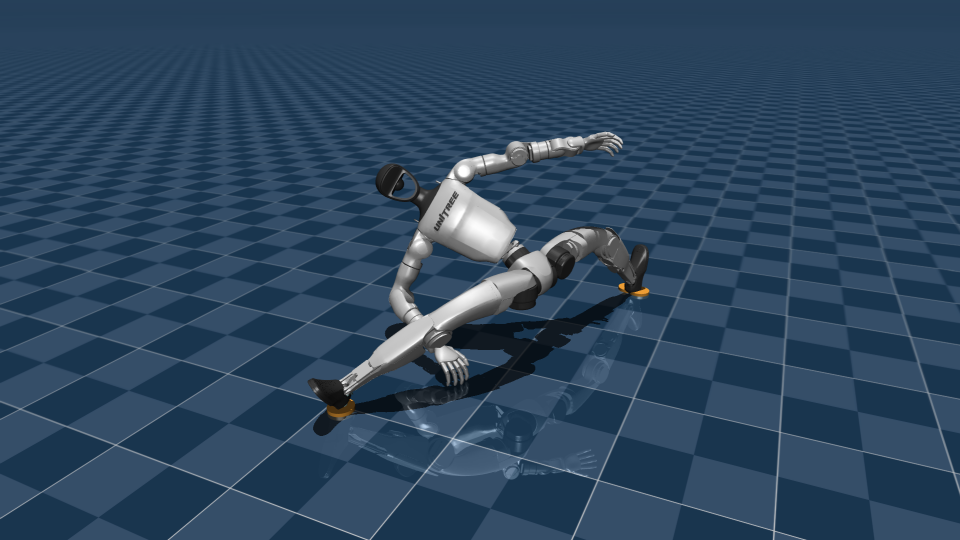}{Leg-Swing}
    \end{minipage}

    \vspace{2pt}

    \begin{minipage}[t]{0.48\textwidth}\centering
      \humimg{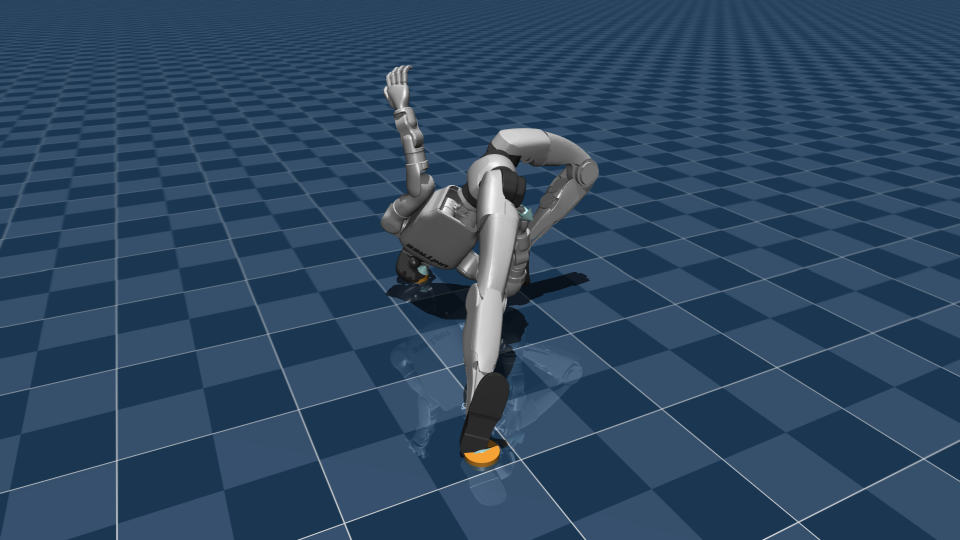}{Lateral Roll}
    \end{minipage}\hfill
    \begin{minipage}[t]{0.48\textwidth}\centering
      \humimg{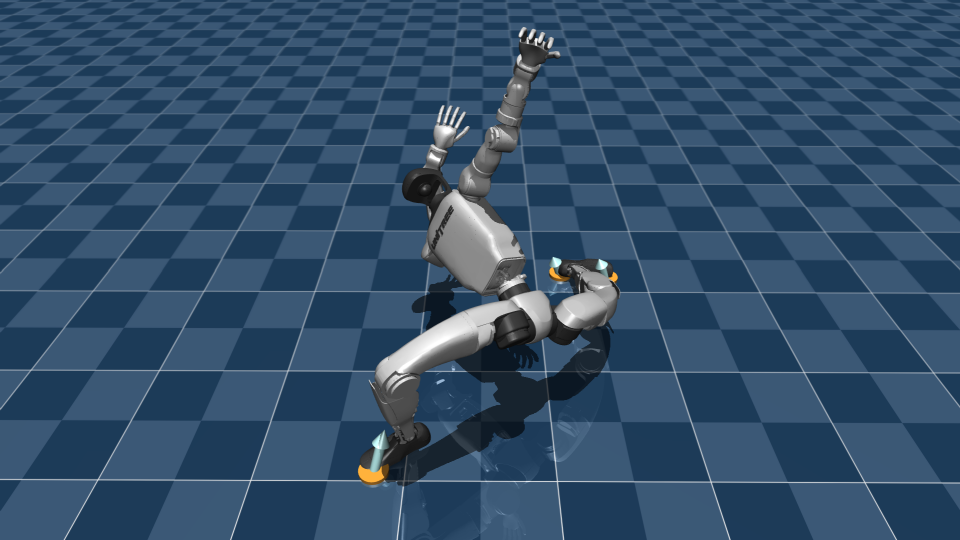}{Kip-Up}
    \end{minipage}

    \vspace{1pt}
    {\footnotesize (b) Humanoid standup}
  \end{minipage}

  \vspace{4pt}
  \captionof{figure}{Multi-modal trajectory modes discovered by MToD
    on two tasks with qualitatively distinct solution families.
    \textbf{(a)}~Pendulum swing-up ($0$--$3.5$\,s): two
    globally optimal modes---Right Swing~(top) and Left
    Swing~(bottom)---that are mirror-symmetric about the pivot
    axis (dashed line).
    Orange arrows indicate tip velocity.
    \textbf{(b)}~Humanoid standup (29-DOF Unitree~G1): four
    modes arising from distinct contact sequences and
    whole-body strategies.
    Orange spheres mark active floor contacts and colored arrows
    show contact force vectors.
    Both tasks are inherently multi-modal: the pendulum admits
    two symmetric near-global optima, while the humanoid exhibits
    a rich family of locally optimal strategies distinguished
    by their contact patterns.}
  \label{fig:teaser}
\end{center}

%% file: sections/abstract.tex
\begin{abstract}
Trajectory optimization is fundamental to robotics and model predictive control (MPC), but in contact-rich or nonconvex settings, it often behaves as a local method whose outcome is dominated by initialization. Moreover, many tasks are intrinsically multi-modal, admitting several qualitatively distinct high-quality solution families that single-trajectory optimizers or unimodal sampling distributions tend to miss. We propose \emph{Multi-Modal Trajectory optimization via Diffusion} (MToD), a training-free framework that casts trajectory optimization as sampling from a Boltzmann distribution over control sequences and solves it using a denoising-style reverse diffusion process. Rather than learning a denoiser or differentiating through dynamics, MToD estimates diffusion scores via Monte Carlo importance weighting from physics-simulator rollouts, enabling coarse-to-fine global exploration while naturally preserving multiple modes. Building on this sampler, \emph{MToD warm-started MPC} decouples offline multi-modal discovery from online execution: diverse candidate trajectories are generated and refined offline, compact mode sets are extracted for warm starts, and a lightweight tracker selects and tracks the modes in closed loop. We evaluate the methods on a numerical benchmark, a torque-limited pendulum swing-up, and a high-dimensional humanoid standup task. Across all three settings, our methods consistently discover multiple qualitatively distinct trajectory modes and achieve lower closed-loop cost than representative baselines.
\end{abstract}

%% file: sections/introduction.tex
\section{Introduction}
Trajectory optimization is a cornerstone of robotics and control, underpinning model predictive control, motion planning, and contact-aware manipulation \cite{Posa2014ContactTO, AceitunoCabezas2020QuasiDynamic}. Nevertheless, because these problems are often highly nonconvex---with hard constraints, hybrid dynamics, and discontinuous contact events---most practical solvers behave as \emph{local} optimizers in disguise: they refine what they are given. As a result, the solution quality is frequently dominated by the basin of attraction induced by the initial trajectory (or the initial sampling distribution), so an unfavorable warm start can lead to infeasible rollouts, premature convergence, or systematically suboptimal behaviors. Consequently, while modern trajectory optimizers can be very effective once placed near a good basin, endowing them with reliable \emph{global exploration} remains a central and unresolved challenge.

Beyond nonconvexity per se, many robotics tasks are inherently \emph{multi-modal} in the sense that high-quality solutions form several qualitatively distinct clusters, rather than a single unimodal neighborhood. This structure is often dictated by task topology and environment geometry: for instance, obstacle avoidance naturally induces different homotopy classes (e.g., passing on the left versus the right), each corresponding to a distinct mode of the trajectory distribution \cite{Bhattacharya2010Homotopy}. Moreover, the need for \emph{multi-modal} reasoning becomes particularly acute in contact-rich domains, where combinatorially many contact sequences and frictional interactions yield a large family of feasible strategies \cite{Posa2014ContactTO}. In such settings, it is typically insufficient to merely refine a single nominal plan; instead, discovering multiple stable contact modes---each representing a coherent contact pattern coupled with an associated motion strategy---is often prerequisite for robustness and high performance, especially under model mismatch and external perturbations.

With the above two challenges in mind, one natural direction is to resort to global trajectory optimization in order to improve \emph{global exploration} and approach global or near-global optimality \cite{kang2024strom}. In practice, however, such guarantees often come at the cost of restrictive problem structure or modeling assumptions---for example, semidefinite/sum-of-square relaxations typically require polynomial objectives and constraints \cite{kang2024strom}, while mixed-integer convex or quasi-dynamic contact formulations rely on carefully chosen relaxations and approximations of hybrid contact dynamics \cite{AceitunoCabezas2020QuasiDynamic,pang2023globalsmoothing}. Moreover, because these solvers are usually designed to return a \emph{single} (near-)optimal solution, they do not, by default, preserve the \emph{multi-modality} of the solution landscape; recovering a \emph{set} of diverse, high-quality modes often requires repeated solves or additional diversity mechanisms \cite{Marcucci2024GCS}.

Alternatively, learning-based approaches can amortize exploration across experience, but they trade away the on-demand, training-free nature of trajectory optimization; this includes diffusion-based controllers that learn a generative model of optimal control sequences offline \cite{huang2025diffusionnmpc}. In particular, standard reinforcement learning pipelines ultimately produce a single policy (or a single action distribution) and, unless explicitly regularized for diversity, may under-represent distinct solution modes despite extensive exploration during training. Imitation- and dataset-driven methods face complementary limitations: obtaining demonstrations that are both high-quality and genuinely multi-modal is costly \cite{Ross2011DAgger}, and can be particularly challenging in high-DoF contact-rich systems (e.g., humanoids) and fast, hybrid, underactuated regimes (e.g., inverted pendulum swing-up), where rare yet critical modes and transitions are difficult to capture and curate.

The above considerations motivate a fundamental question: \emph{can we endow trajectory optimization with a training-free mechanism for global exploration that systematically discovers a set of diverse, high-quality solution modes?}

To address this question, we propose \textbf{Multi-Modal Trajectory Optimization via Diffusion (MToD)} to repurpose diffusion as a principled annealed sampling mechanism operating directly in control space---without any offline training, learned score networks, or demonstration data. Specifically, the key insight is that diffusion's coarse-to-fine noise schedule provides a natural curriculum for global exploration: at high noise levels the search is broad enough to reach distant basins of attraction, whereas successive denoising steps progressively concentrate samples around high-quality trajectories. Finally, by adopting a pipeline that decouples offline mode discovery from online execution, we translate this multi-modal sampling capability into real-time closed-loop control. To summarize, our contributions are threefold:
\begin{enumerate}
  \item \textbf{MToD: training-free diffusion for trajectory
    optimization.} A reverse-diffusion sampler that discovers
    multiple trajectory modes directly from physics-simulator
    rollouts, without any offline training or demonstration data.
  \item \textbf{MToD warm-started MPC.} A pipeline that decouples
    offline multi-modal exploration from online real-time
    execution, enabling mode-conditioned warm-starting for
    closed-loop control.
  \item \textbf{Robotic validation.} We demonstrate reliable
    multi-modal discovery and up to $39\%$ cost reduction over the
    standard baseline on a pendulum swing-up and a 29-DOF humanoid
    standup task.
\end{enumerate}

%% file: sections/related_work.tex
\section{Related Work}

\subsection{Trajectory Optimization with Global Exploration}
Many trajectory optimizers and sampling-based MPC methods address nonconvexity via stochastic \emph{global exploration}, including information-theoretic MPC \cite{williams2018itmcp} and MPPI-style controllers and variants \cite{crestaz2025td, alvarez2025real} that iteratively reweight samples to update a proposal distribution. In parallel, \emph{global} trajectory optimization pursues stronger guarantees by exploiting structure, e.g., certifiable relaxations \cite{kang2024strom}, hybrid graph-based formulations such as graphs of convex sets \cite{Marcucci2024GCS}, and contact-rich planning enabled by quasi-dynamic modeling and smoothing \cite{pang2023globalsmoothing, AceitunoCabezas2020QuasiDynamic, Posa2014ContactTO}. A complementary direction encourages diverse solutions through discrete search structure, such as homotopy-class constraints \cite{Bhattacharya2010Homotopy} or globally structured motion planning schemes \cite{le2025global}. 
Compared to these approaches, our method explicitly targets multi-modal trajectory discovery while still seeking the globally best solution.

\subsection{Diffusion Models for Planning and Control}
Diffusion models, such as DDPMs \cite{ho2020ddpm}, generate samples via iterative denoising and have recently become a versatile tool for robotics planning and control. A dominant direction treats diffusion as a learned policy class trained from demonstrations or offline data, exemplified by diffusion policy \cite{chi2023diffusionpolicy} and other amortized generative controllers for fast online execution \cite{kurtz2025gpc}. Closely related, diffusion has also been used as a learned generative prior for motion planning  \cite{carvalho2023motion}, enabling conditional generation of robot motions and plans. A related planning line \cite{huang2025diffusionnmpc, julbe2025diffusion} learns a generative model of (near-)optimal action or control sequences offline (e.g., to imitate an MPC solver or to generalize across initializations). 
Complementary model-based approaches \cite{pan2024model, xue2024dialmpc, kurtz2025equality} reduce reliance on demonstrations by leveraging known dynamics and rollouts, using diffusion-style annealing or sampling to improve exploration and robustness in trajectory optimization and MPC. 
In contrast to prior diffusion-for-control works, we use diffusion as a training-free annealed sampler directly in control space to simultaneously recover multiple qualitatively distinct trajectory modes, and identify the globally best trajectory among these modes.

%% file: sections/methodology.tex
\section{Preliminaries}
\label{sec:prelim}

\subsection{Problem Formulation}
\label{sec:prelim:direct}

We consider a discrete-time system with state $\bx_t\in\mathbb{R}^{n_x}$,
control $\bu_t\in\mathbb{R}^{n_u}$, and dynamics $\bx_{t+1}=f(\bx_t,\bu_t)$
over a horizon of $T$ steps. In direct transcription, both states and
controls are decision variables and dynamics are enforced as equality
constraints:
\begin{subequations}\label{eq:direct}
\begin{align}
  \min_{\bx_{0:T},\,\bu_{0:T-1}} \;\;
    & \ell_T(\bx_T) + \textstyle\sum_{t=0}^{T-1}\ell_t(\bx_t,\bu_t)
      \label{eq:direct:cost}\\
  \text{s.t.}\;\;
    & \bx_0 = \bx_{\mathrm{init}},
      \label{eq:direct:ic}\\
    & \bx_{t+1}=f(\bx_t,\bu_t), \;\; t=0,\dots,T{-}1,
      \label{eq:direct:dyn}\\
    & \bg(\bx_t,\bu_t)\le \mathbf{0}, \;\; t=0,\dots,T{-}1,
      \label{eq:direct:pathcon}\\
    & \bg_T(\bx_T)\le \mathbf{0}.
      \label{eq:direct:termcon}
\end{align}
\end{subequations}
Here $\bg$ and $\bg_T$ encode path and terminal constraints, and
$\ell_t,\ell_T$ are stage and terminal costs. This optimization problem is challenging in
robotics because (i) $f$ is typically nonlinear and underactuated, (ii)
constraints such as collision avoidance are
nonconvex, and (iii) contact-rich tasks further introduce hybrid,
nonsmooth dynamics and feasibility conditions. Moreover, the resulting
objective landscape is often \emph{multi-modal}: distinct homotopy
classes or contact sequences correspond to different locally optimal
basins. As a consequence, conventional solvers and many sampling-based
optimizers tend to converge to a single basin dictated by initialization
or by a progressively tightening sampling distribution.
These properties motivate our training-free, diffusion-based multi-modal trajectory optimizer.

\subsection{Denoising Diffusion Probabilistic Models}
\label{sec:prelim:ddpm}

Denoising diffusion probabilistic models (DDPMs) define a
variance-preserving (VP) forward noising process
$p_0 \to p_1 \to \cdots \to p_N$ and a reverse denoising process that
maps samples from $p_N \approx \mathcal{N}(\mathbf{0},\bI)$ back toward
$p_0$ \cite{ho2020ddpm}. Given $\bU^{(0)} \sim p_0$, the forward Markov
chain is, for $i=1,\ldots,N$,
\begin{equation}\label{eq:ddpm_forward}
  q(\bU^{(i)} \mid \bU^{(i-1)})
  = \mathcal{N}\!\bigl(\sqrt{\alpha_i}\,\bU^{(i-1)},\; \beta_i \bI\bigr),
\end{equation}
where $\beta_i\in(0,1)$ is the noise schedule, $\alpha_i \coloneqq
1-\beta_i$, and $\bar{\alpha}_i \coloneqq \prod_{k=1}^{i}\alpha_k$ (with
$\bar{\alpha}_0 \coloneqq 1$). This yields the closed-form marginal
\begin{equation}\label{eq:forward}
  q(\bU^{(i)} \mid \bU^{(0)})
  = \mathcal{N}\!\bigl(\sqrt{\bar{\alpha}_i}\,\bU^{(0)},\;
  (1-\bar{\alpha}_i)\bI\bigr),
\end{equation}
and for typical schedules with $\beta_1=\beta_{\min}$ and
$\beta_N=\beta_{\max}$, one obtains $p_N \approx \mathcal{N}(\mathbf{0},\bI)$.
The reverse process is modeled as a Gaussian transition, for
$i=N,\ldots,1$,
\begin{equation}\label{eq:ddpm_reverse_gauss}
  p_\theta(\bU^{(i-1)} \mid \bU^{(i)})
  = \mathcal{N}\!\bigl(\mu_\theta(\bU^{(i)}, i),\; \tilde{\beta}_i \bI\bigr),
\end{equation}
with VP posterior variance
\begin{equation}\label{eq:ddpm_tildebeta}
  \tilde{\beta}_i
  \coloneqq \beta_i\,\frac{1-\bar{\alpha}_{i-1}}{1-\bar{\alpha}_i}.
\end{equation}
Equivalently, the reverse transition can be written in a score-based
form as, for $i=N,\ldots,1,\, \bz\sim\mathcal{N}(\mathbf{0},\bI)$, 
\begin{equation}\label{eq:ddpm_reverse_score_prelim}
  \bU^{(i-1)}
  = \frac{1}{\sqrt{\alpha_i}}\Bigl(
      \bU^{(i)} + \beta_i\,\nabla_{\bU^{(i)}}\log p_i(\bU^{(i)})
    \Bigr)
  + \sqrt{\tilde{\beta}_i}\bz,
\end{equation}
where $p_i$ denotes the marginal at noise level $i$. In standard DDPMs,
the mean $\mu_\theta$ (equivalently, a noise predictor) is parameterized
by a neural network trained on data \cite{ho2020ddpm}. Our method adopts
the same VP forward--reverse structure, but replaces the learned
denoiser with a training-free score estimator computed from physics
rollouts.

\section{Methodology}
\label{sec:method}
\subsection{Multi-Modal Trajectory Optimization via Diffusion}
\label{sec:method:mtop}

We eliminate the state decision variables in~\eqref{eq:direct} by adopting
a single-shooting parameterization: the dynamics constraints induce a
forward rollout map $\bx_{0:T}=\Phi(\bU)$ from controls to states, where
$\bU\coloneqq \bu_{0:T-1}\in\mathbb{R}^{Tn_u}$. In practice, we
parameterize $\bU$ with $N_k\ll T$ knot points
$\bc=[c_1,\ldots,c_{N_k}]\in\mathbb{R}^{N_k n_u}$ and interpolate to
obtain the full-horizon control sequence. Let $\mathcal{I}(\bc)\in
\mathbb{R}^{Tn_u}$ denote the interpolation operator. We perform rollouts
using $\mathcal{I}(\bc)$ and abuse notation by writing $\bU$ for the
resulting control variable.

\noindent\textbf{Sampling as Optimization.}
We recast trajectory optimization as sampling from a Boltzmann
distribution over control sequences,
\begin{equation}\label{eq:boltzmann}
  p_0(\bU) \propto \exp\!\Bigl(-\frac{J(\bU)}{\lambda}\Bigr),
\end{equation}
where $J(\bU)$ is evaluated by forward rollout and $\lambda>0$ is a
temperature parameter. Dynamics feasibility is satisfied by construction
through $\bx_{0:T}=\Phi(\bU)$. Other state- and task-dependent
constraints are enforced during sampling by assigning zero weight to
infeasible rollouts. In both the reverse-diffusion stage and the MCSA
refinement stage, control bounds are handled directly at the proposal
level. The proposal factorizes over knot indices and control dimensions,
and samples each coordinate independently from a one-dimensional
truncated Gaussian via the inverse-CDF transform
\begin{equation}\label{eq:trunc_gauss}
  u=\mu+\sigma\,\Phi^{-1}\!\bigl(\Phi(a)+(\Phi(b)-\Phi(a))\,v\bigr),
  \quad v\sim\mathcal{U}(0,1),
\end{equation}
where $a=(u_{\min}-\mu)/\sigma$ and $b=(u_{\max}-\mu)/\sigma$. This
ensures the sampled knots respect actuator limits without using projection
as the primary proposal mechanism. We also defensively clip every returned
knot tensor. Because cubic interpolation can overshoot between feasible
knots, every dense control sent to the dynamics is clipped once more:
\begin{equation}\label{eq:dense_control_clip}
  \bu_t^{\mathrm{exec}}
  = \operatorname{clip}\!\left([\mathcal{I}(\bc)]_t,
      \bu_{\min},\bu_{\max}\right).
\end{equation}
Thus proposal truncation enforces feasibility in the search space, while
Eq.~\eqref{eq:dense_control_clip} enforces feasibility of the physically
executed, interpolated control. Since direct sampling from $p_0$ is
intractable in high dimensions, we employ a DDPM-style reverse diffusion
driven by the score $\nabla_{\bU^{(i)}}\log p_i(\bU^{(i)})$; the
corresponding score-based reverse update is given in
Eq.~\eqref{eq:ddpm_reverse_score_prelim}.

\noindent\textbf{Monte Carlo Score Estimation.}
To execute the reverse diffusion in Eq.~\eqref{eq:ddpm_reverse_score_prelim}
without training a denoiser network, we require a training-free
estimate of the score $\nabla_{\bU^{(i)}}\log p_i(\bU^{(i)})$.
Let $p_i$ denote the marginal distribution at noise level $i$ induced by
the VP forward process~\eqref{eq:forward}. By Tweedie-style identities~\cite{Efron2011Tweedie},
the score admits the posterior-mean form
\begin{equation}\label{eq:score}
  \nabla_{\bU^{(i)}} \log p_i(\bU^{(i)})
  = \frac{-\bU^{(i)} + \sqrt{\bar{\alpha}_i}\,
  \mathbb{E}[\bU^{(0)}\mid \bU^{(i)}]}
  {1-\bar{\alpha}_i},
\end{equation}
so it suffices to estimate $\mathbb{E}[\bU^{(0)}\mid \bU^{(i)}]$.

We approximate this posterior mean by importance sampling. Specifically,
we draw $S$ candidates $\bU_j^{(0)}$ from an element-wise box-truncated
Gaussian proposal
\begin{equation}\label{eq:proposal}
\begin{aligned}
  \varphi_i^{\mathrm{box}}(\bU^{(0)};\bU^{(i)})
  &= \prod_{k=1}^{N_k}\prod_{d=1}^{n_u}
    \mathcal{TN}\!\left(
      U_{k,d}^{(0)};\,\mu_{i,k,d},\sigma_{\mathrm{prop}}^2,
      u_{\min,d},u_{\max,d}\right), \\
  \mu_i &= \frac{\bU^{(i)}}{\sqrt{\bar{\alpha}_i}}, \\
  \sigma_{\mathrm{prop}}^2
  &= \frac{1-\bar{\alpha}_i}{\bar{\alpha}_i},
\end{aligned}
\end{equation}
where $\mathcal{TN}(\cdot;\mu,\sigma^2,l,h)$ denotes a one-dimensional
Gaussian conditioned on $[l,h]$ and is sampled using
Eq.~\eqref{eq:trunc_gauss}. The untruncated center and scale are obtained
by viewing the DDPM forward kernel
$q(\bU^{(i)}\mid\bU^{(i-1)})$ in Eq.~\eqref{eq:ddpm_forward} as a
distribution over the clean control $\bU^{(0)}$ when the noisy iterate
$\bU^{(i)}$ is fixed. They coincide with the Gaussian proposal used
in~\cite{pan2024model}; conditioning each coordinate on its task-specific
actuator interval is our constraint-handling layer.

Rolling out each candidate yields its cost $J(\bU_j^{(0)})$ and we assign the unnormalized Boltzmann weight
\begin{equation}\label{eq:unnorm_weights}
  \tilde{w}_j = \exp\!\Bigl(-\frac{J(\bU_j^{(0)})}{\lambda}\Bigr),
\end{equation}
set $\tilde{w}_j=0$ for infeasible rollouts, and normalize
$w_j=\tilde{w}_j/(\sum_{k=1}^{S}\tilde{w}_k+\epsilon)$, where
$\epsilon>0$ prevents division by zero if all samples are infeasible.
In practice, we compute the normalization with a log-sum-exp routine to
avoid numerical underflow. The posterior mean is then approximated by
\begin{equation}\label{eq:post_mean_approx}
    \mathbb{E}[\bU^{(0)}\mid \bU^{(i)}] \approx \sum_{j=1}^{S} w_j\,\bU_j^{(0)},
\end{equation}
which yields the score via~\eqref{eq:score}. One diffusion chain requires $\mathcal{O}(NS)$ rollouts. Running $K$ chains increases the cost to $\mathcal{O}(KNS)$ rollouts, but the chains can be evaluated in parallel.

\begin{figure}[htp]
  \centering
  \includegraphics[width=0.5\textwidth]{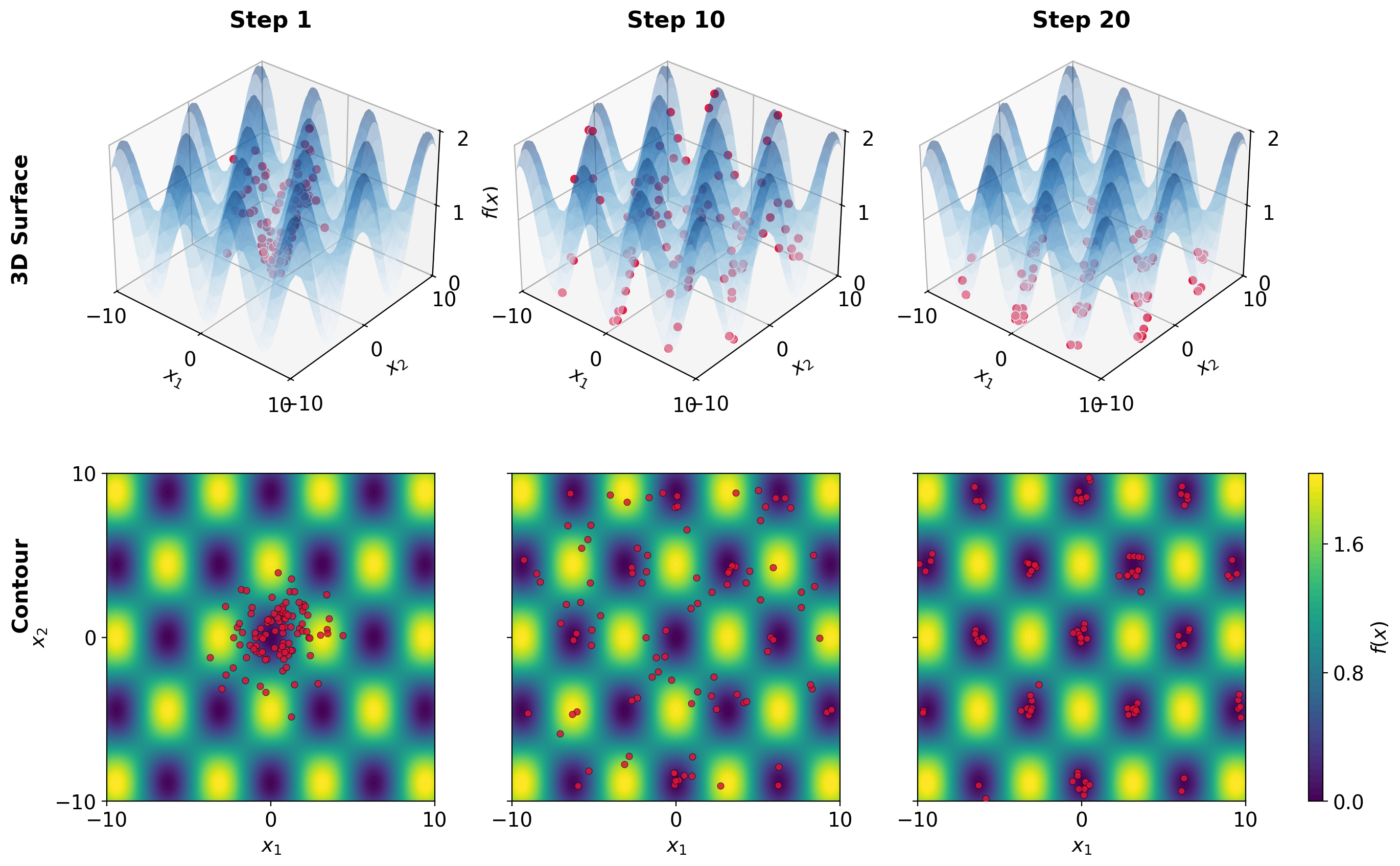}
  \caption{\textbf{MToD global exploration on the Griewank function}
  (global minimum at the origin, $f(\mathbf{0}){=}0$).
  Snapshots of one MToD chain at steps 1, 10, and 20 (top: 3D surface;
  bottom: contour view). Samples transition from broad exploration to
  concentration around multiple low-cost basins. Parameters: $N{=}20$
  steps, $\lambda{=}0.1$, $S{=}2048$ samples/step, linear schedule
  $\beta_{\min}{=}10^{-4}$ to $\beta_{\max}{=}0.5$.}
  \label{fig:griewank_ddpm}
\end{figure}

\noindent\textbf{Global Exploration and Multi-Modality.}
As $\lambda \to 0$, the Boltzmann target $p_0$ concentrates on globally
optimal solutions; for finite $\lambda$, it retains probability mass
across multiple basins and thus explicitly represents multi-modality.
MToD leverages this structure through the \emph{stochastic} DDPM reverse
process: early steps inject substantial noise (large
$\tilde{\beta}_i$), encouraging basin hopping and global coverage,
whereas late steps suppress noise and refine within each basin.
Fig.~\ref{fig:griewank_ddpm} illustrates this coarse-to-fine behavior
on the 2D Griewank function: across parallel chains, MToD attains near-global objective values ($1.3{\times}10^{-3}$ above the optimum) and
covers all $17$ low-cost local minima ($f{<}0.1$) within a proximity
radius of~$0.5$, confirming that stochastic reverse diffusion maintains diversity
during denoising rather than collapsing to a single solution.

\subsection{MToD Warm-Started MPC}
\label{sec:method:warm_mpc}

MToD separates offline mode discovery from online execution in four
phases.
First, MToD launches $K$ independent DDPM chains in parallel on GPU,
each executing $N$ reverse-diffusion steps with $S$ rollout samples; the stochastic reverse process
causes different chains to settle in different cost basins.
Second, each endpoint is refined by the deterministic score ascent
updates:
\begin{equation}\label{eq:mcsa}
  \bU^{(i-1)}
  = \frac{\sqrt{\bar{\alpha}_i}}
         {\sqrt{\alpha_i}}
    \mathbb{E}[\bU^{(0)}\mid \bU^{(i)}],
  \quad
\end{equation}
where the expectation is approximate via the Monte Carlo method (\ref{eq:post_mean_approx}).
This equation ascends the score because the posterior-mean denoising term is proportional to $\nabla_{U^{(i)}} \log p(U^{(i)})$, so the update moves the sample toward higher-density regions.
The Monte Carlo candidates used for every MCSA update are drawn from the
same factorized, element-wise truncated proposal in
Eq.~\eqref{eq:proposal}; Stage~2 therefore preserves the proposal-level
actuator constraints used in Stage~1.
Third, MToD forward-simulates each refined trajectory to obtain a
task-specific behavioral feature embedding, and selects modes
by greedy diversity filtering in the normalized behavioral space:
candidates are processed in ascending cost order and accepted only
if their distance to all previously selected modes exceeds a merge
radius~$r$, up to a budget of~$M$ modes.
Finally, each extracted mode serves as a reference trajectory for an
MPPI controller~\cite{williams2018itmcp} that takes the MToD
reference into its sampling mean during an initial window.

\subsection{Task-Specific Mode Extraction}
\label{sec:method:bpca}

The behavioral embedding in the third phase must capture
task-relevant distinctions between solution modes. Its design
depends on the structure of the multi-modality present in each task.

\noindent\textbf{Pendulum swing-up.}
The torque-limited pendulum admits exactly two topologically distinct
swing-up families---right swing ($\theta > 0$) and left swing
($\theta < 0$)---that reach the same upright equilibrium via mirrored
trajectories. Because the multi-modality reduces to a single binary
degree of freedom, the committed swing direction, we bypass a general
feature embedding and instead apply a direct \emph{swing classifier}:
each chain is classified by the sign of its final angle $\theta_k(T)$,
and the lowest-cost representative from each group is selected, yielding
$M{=}2$ modes. We use the final rather than the initial angle
because the pendulum often employs energy pumping, whose initial motion opposes the committed swing direction.

\noindent\textbf{Humanoid standup.}
The 29-DOF humanoid exhibits rich, high-dimensional multi-modality
driven by combinatorially many feasible contact sequences. Unlike the pendulum, no simple physical classifier can distinguish standup strategies, so we construct a 21-dimensional behavioral feature
vector from each forward-simulated trajectory. The features capture
per-limb contact fractions and switch counts, feet-only vs.\
hand-involved support ratios, torso height dynamics including minimum, maximum, final height, and time to half-height, CoM kinematics such as peak horizontal speed, mean vertical velocity, and net drift, as well as left--right hand asymmetry. After z-score normalization, Principal Component Analysis (PCA) decorrelates the features so that the subsequent Euclidean distance metric weights independent behavioral axes equally, rather than double-counting correlated groups. Greedy radius-based selection then identifies $M{=}4$ behaviorally distinct standup strategies while preserving cost ordering.

%% file: sections/exp_gmm.tex
\subsection{Gaussian Mixture Function Benchmark\protect\footnote{We will open-source the code upon publication}}
\label{sec:exp:gmm}

\noindent\textbf{Objective Function.}
To evaluate MToD on a controlled multi-modal landscape with known
ground truth, we adopt the two-dimensional Gaussian Mixture Model (GMM)
benchmark. The objective function is 
defined as the negated mixture density plus a quadratic regularizer:
\begin{equation}\label{eq:gmm_obj}
  F(\bx) \;=\; -p(\bx) \;+\; c_{\mathrm{reg}}\,\lVert \bx - \bx_{\mathrm{reg}}
  \rVert^2,
\end{equation}
where $p(\bx)$ is a mixture of $25$ isotropic Gaussians arranged on
a $5\times 5$ grid,
\begin{equation}\label{eq:gmm_density}
  p(\bx) = \sum_{i=1}^{25}
  \frac{w_i}{2\pi\sigma}\,
  \exp\!\Bigl(-\frac{\lVert \bx-\bm{m}_i\rVert^2}{2\sigma}\Bigr).
\end{equation}
Each mode center $\bm{m}_i$ is a point in the meshgrid
$\{0,1,2,3,4\}^2$, and the covariance of every component is
$\Sigma_i = \sigma\,\bI_2$ with $\sigma = 0.1$. This small variance
produces 25 well-separated basins whose depths are solely governed by
the weight vector $\bm{w} = (w_1,\ldots,w_{25})$. The regularization
term uses $c_{\mathrm{reg}} = 0.005$ and $\bx_{\mathrm{reg}} = (2,\,2)^\top$,
which gently penalizes the domain periphery without altering the local
basin structure. Minimizing $F$ thus requires locating the deepest
basin among 25 local minima whose relative depths vary with $\bm{w}$.

\noindent\textbf{Benchmark Setup.}
We evaluate MPPI \cite{williams2018itmcp}, MBD \cite{pan2024model} and MToD across
$100$ independent problem instances. For each instance, the
weight vector $\bm{w}$ is drawn from a symmetric Dirichlet
distribution $\mathrm{Dir}(\bm{1}_{25})$ with a distinct seed, yielding
diverse landscapes with varying numbers and depths of dominant modes.
The starting point $\bx_0$ cycles through the four domain corners
$\{(-1,-1),\;(-1,5),\;(5,-1),\;(5,5)\}$ in round-robin fashion, so
that on average every algorithm faces the same distribution of initial
distances to the global optimum. To ensure a fair per-chain comparison, all three algorithms use
identical rollout budgets: $S{=}1024$ samples per step and
$N{=}20$ optimization steps, yielding $S \times N = 20{,}480$
total rollouts per chain. Because MPPI and MBD are deterministic
given a fixed seed, each runs a single chain per instance. MToD,
being a stochastic sampler, runs $K{=}30$ independent chains per
instance (total $30 \times 20{,}480 = 614{,}400$ rollouts); its
best-chain result is used for the global-optimality metric, while
all $K$ chains contribute to the multi-modality analysis.

\noindent\textbf{Metrics.}
We report two complementary metrics. (i)~\emph{Global-optimality gap}:
the residual $\Delta = F(\bx_{\mathrm{final}}) - F(\bx^{*})$, where
$\bx^{*}$ is the analytically known global minimizer for each instance.
For MToD, $\bx_{\mathrm{final}}$ is the best endpoint across $K$
chains. Lower $\Delta$ indicates better convergence to the true global
minimum. (ii)~\emph{Mode capture rate} (MToD only): for a threshold
$X\%$, a mode $i$ is \emph{qualifying} if $F(\bm{m}_i) \le
F(\bx^{*}) + \tfrac{X}{100}\,(F_{\mathrm{worst}} - F(\bx^{*}))$,
where $F_{\mathrm{worst}}$ is the highest-valued mode center. A
qualifying mode is \emph{captured} if any MToD final point lies within
$\ell_2$ distance $0.3$ of $\bm{m}_i$. The capture rate is the fraction
of qualifying modes captured, averaged over 100~instances.

\begin{figure}[t]
  \centering
  \includegraphics[width=0.49\columnwidth]{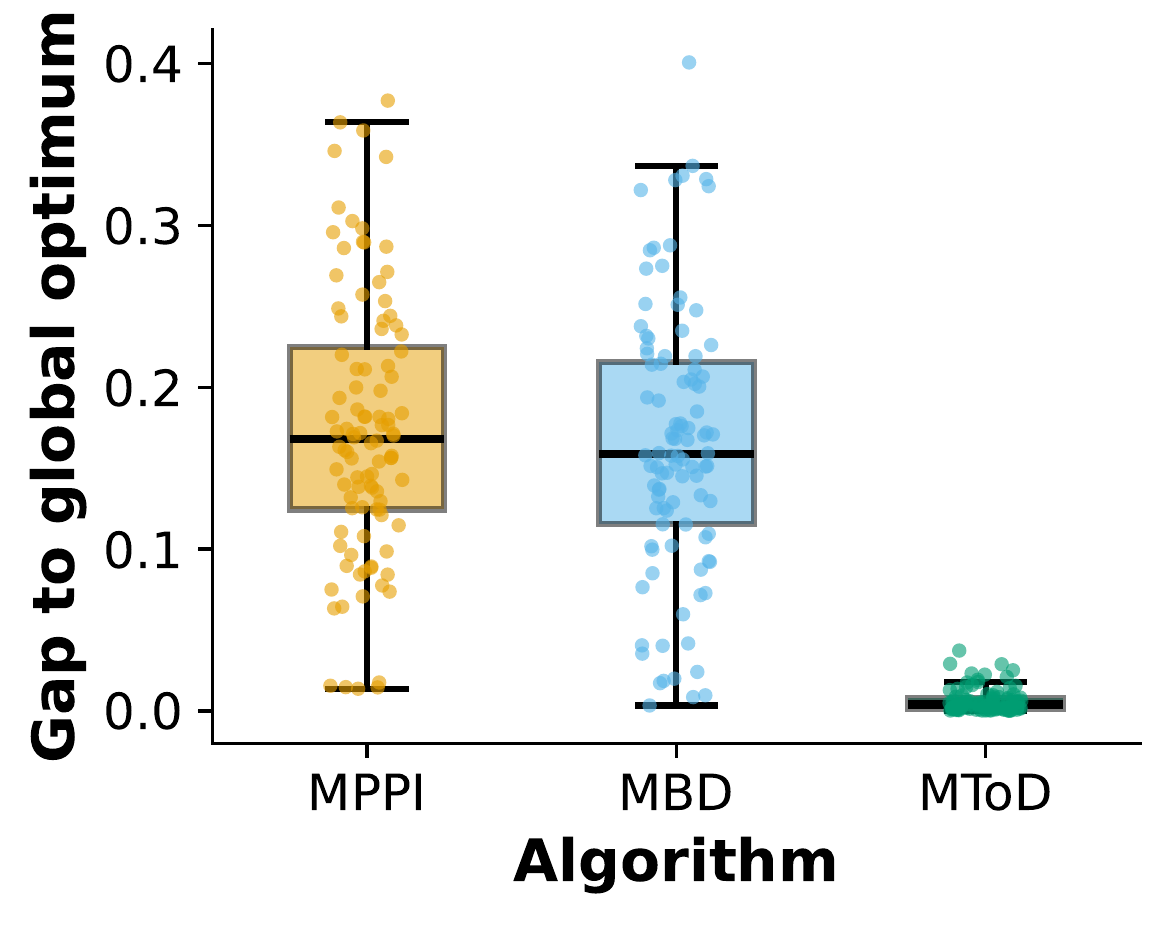}%
  \hfill
  \includegraphics[width=0.49\columnwidth]{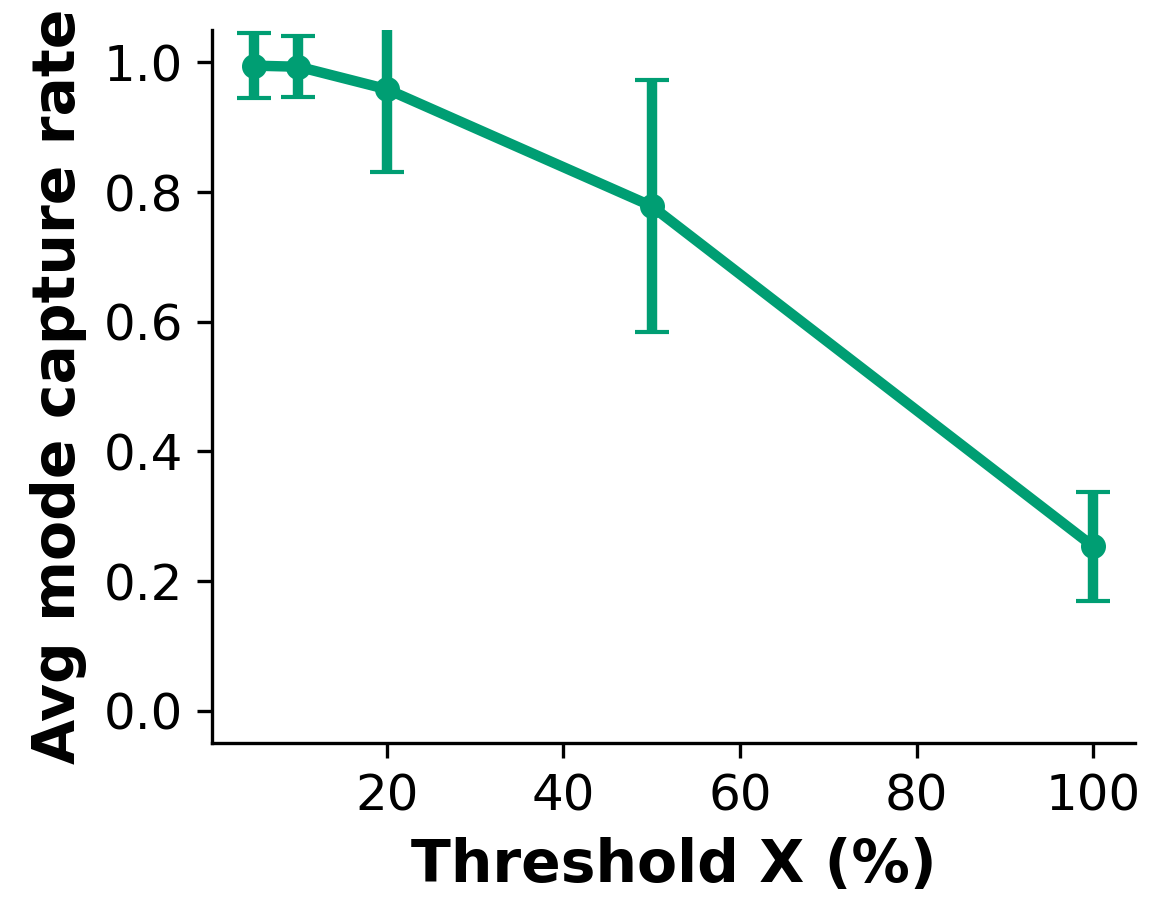}
  \caption{GMM benchmark results over 100 randomized instances.
    \emph{Left:} Global-optimality gap distribution for MPPI, MBD, and
    MToD (box-and-jitter plot; lower is better).
    \emph{Right:} MToD mode capture rate as a function of qualifying
    threshold $X\%$ (mean $\pm$ std across instances; higher is
    better).}
  \label{fig:gmm_results}
\end{figure}

\noindent\textbf{Results.}
Fig.~\ref{fig:gmm_results} summarizes both metrics over 100~instances.
The box plot (left) shows that MToD's global-optimality gap is tightly
concentrated near zero: its mean gap is $6.24{\times}10^{-3}$
(median $3.86{\times}10^{-3}$), approximately $28\times$ smaller than
MPPI ($0.174$) and $26\times$ smaller than MBD ($0.165$). Even in the
worst case, MToD's gap ($0.037$) remains an order of magnitude below
the average gaps of both baselines. These results demonstrate that the
stochastic reverse-diffusion process enables MToD to escape local
basins that trap single-trajectory optimizers, consistently converging
near the global minimum despite randomized landscapes and distant
initialization.

The mode-capture curve (right) shows that MToD maintains near-perfect
capture for the most critical modes: at tight thresholds
($X{\le}20\%$), where only the top $1$--$2$ modes qualify, capture
exceeds $95\%$. As the threshold widens to $X{=}50\%$, roughly $5$
modes qualify per instance and MToD still covers $77.8\%$. At
$X{=}100\%$, all ${\sim}25$ modes qualify and the 30-chain budget
captures $6.2$ modes on average ($25.3\%$). This graceful decline
reflects a natural trade-off between chain count $K$ and total mode
count: with 30~chains, MToD reliably identifies the globally best
modes while maintaining substantial coverage of secondary basins---a
property valuable for warm-starting online MPC with diverse candidate
trajectories.

\begin{figure}[t]
  \centering
  \includegraphics[width=\columnwidth]{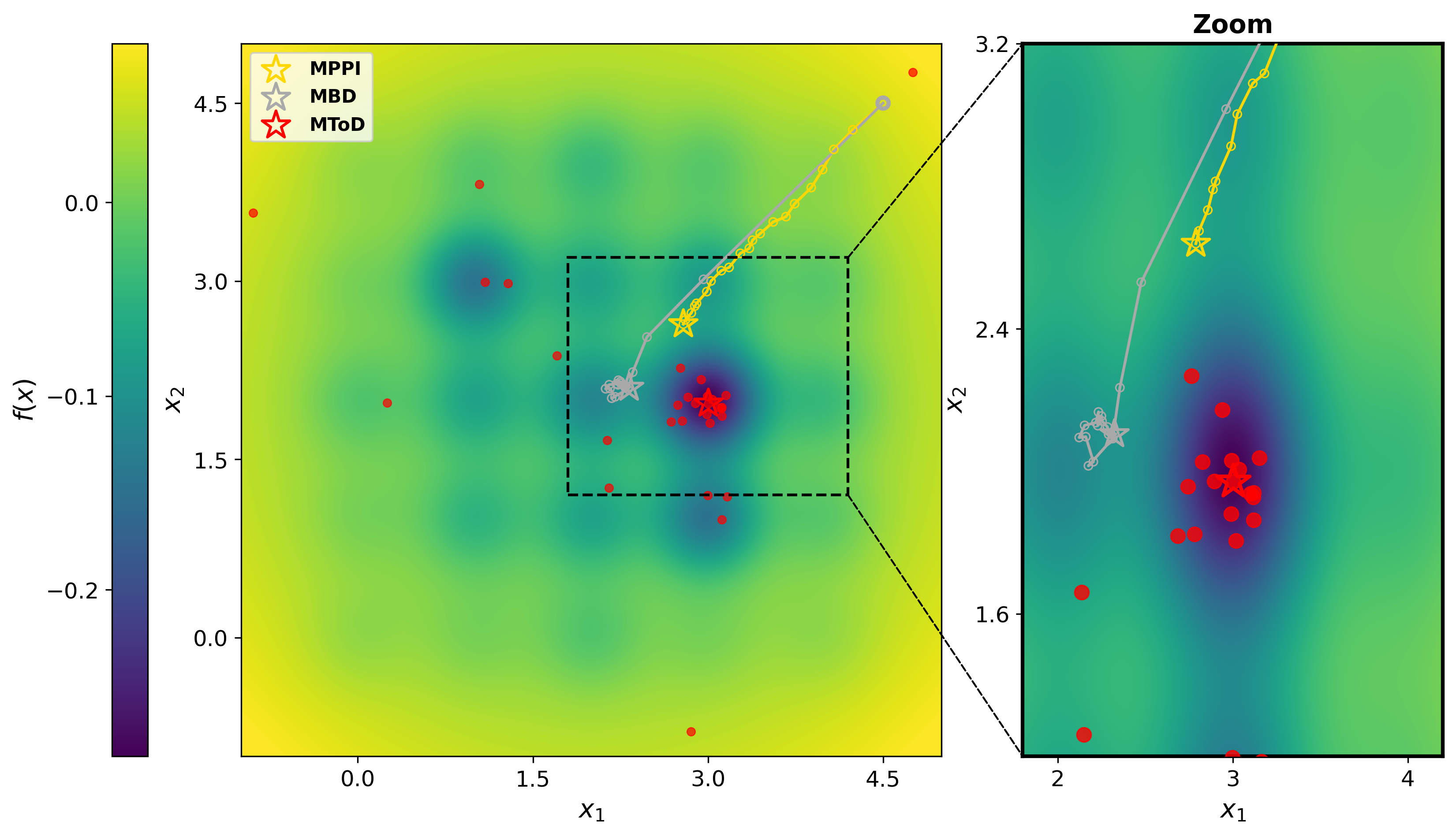}
  \caption{Single-instance GMM comparison. The contour shows the
    objective $F(\bx)$; darker regions indicate deeper basins.
    MPPI (yellow stars) and MBD (gray stars) each follow a single
    descent path from the upper-right corner and converge to
    suboptimal modes.  MToD's 30 chains (red dots) spread across 7
    distinct basins, with the majority concentrating near the global
    minimum at $(3,\,2)$.  The right panel zooms into the dashed
    rectangle, revealing the dense cluster of MToD chains around
    the global basin.}
  \label{fig:gmm_comparison}
\end{figure}

\noindent\textbf{Qualitative Example.}
Fig.~\ref{fig:gmm_comparison} illustrates a representative single
instance with hand-designed weights that produce a clearly dominant
global basin at $\bm{m}^{*}{=}(3,\,2)$, where
$F(\bm{m}^{*}){=}{-}0.285$. The starting point is placed at the
far corner $\bx_0{=}(4.5,\,4.5)$, far from any deep basin.
MPPI descends along a greedy gradient path and settles near
$(2.8,\,2.6)$ with $F{=}{-}0.067$ ($\Delta{=}0.218$), trapped in a
saddle region between the $(3,\,3)$ and $(3,\,2)$ modes. MBD follows
a similar trajectory but converges to $(2.3,\,2.1)$ near the
$(2,\,2)$ mode with $F{=}{-}0.093$ ($\Delta{=}0.193$); while this
basin is deeper than MPPI's endpoint, it is still not the global
optimum. In contrast, MToD's $K{=}30$ chains spread across
$7$ of the $25$ modes, with $17$ chains ($57\%$) landing within
distance $0.5$ of the global minimizer. The best chain reaches
$(3.00,\,1.97)$ with $F{=}{-}0.284$, yielding a gap of only
$\Delta{=}0.001$---more than two orders of magnitude smaller than
either baseline. The scatter of red endpoints in
Fig.~\ref{fig:gmm_comparison} visually confirms that MToD
simultaneously discovers multiple high-quality basins rather than
committing to a single descent direction, which is the key property
enabling diverse trajectory proposals for downstream MPC warm-starting.

%% file: sections/exp_pendulum.tex
\subsection{Pendulum Swing-Up}
\label{sec:exp:pendulum}

\noindent\textbf{Task.}
We consider the classic torque-limited pendulum swing-up, where the
controller must drive a pendulum from its stable downward equilibrium
to the inverted (upright) position and hold it there. The system has a
single revolute joint ($n_u{=}1$) and continuous state
$(\theta,\dot\theta)$.  The control horizon spans $1.0$\,s with
$N_k{=}11$ knots and $S{=}128$~samples per online planning step. Because the
torque limit prevents direct swing-up in a single motion, two
topologically distinct solution families exist: right swing
($\theta > 0$) and left swing ($\theta < 0$). This makes the
pendulum an ideal
test bed for multi-modal trajectory optimization.

\noindent\textbf{Setup.}
We compare MToD against three online controllers, all running in
closed loop for $5$\,s of simulation:
\begin{itemize}
\item \emph{MPPI} \cite{williams2018itmcp}: vanilla Model Predictive Path Integral control with
  128~samples, no warm-start.
\item \emph{DIAL-MPC} \cite{xue2024dialmpc}: diffusion-inspired annealing with
  256~samples and 10~iterations per step.
\item \emph{MBD} \cite{pan2024model}: online Monte Carlo Score Ascent with 256~samples
  and 10~iterations per step.
\end{itemize}
MToD uses the four-phase pipeline: $K{=}32$ DDPM chains ($N{=}50$
diffusion steps each), MBD refinement ($20$ iterations), swing-direction mode
extraction (Section~\ref{sec:method:bpca}), and online MPPI tracking
with the best reference trajectory. The swing classifier produces
exactly $M{=}2$ modes per seed---one right swing, one left swing.
The online phase
uses the same MPPI configuration as the baseline (128~samples,
1~iteration), so any performance difference is attributable solely
to the warm-starting reference.
We evaluate over $10$ independent random seeds with non-overlapping
seed ranges.

\noindent\textbf{Results.}
Table~\ref{tab:benchmark} summarizes the benchmark. MToD achieves a
total cost of $8.79 \pm 0.10$, comparable to DIAL
($8.71 \pm 0.14$) and substantially lower than vanilla MPPI
($10.20 \pm 0.34$). MToD reduces cost by $13.8\%$ compared to MPPI.
Across all 10~seeds, the swing classifier reliably produces two
modes with opposite swing directions, and the best-of-two selection
consistently recovers a near-optimal warm start.

\begin{table}[t]
  \centering
  \caption{Benchmark results (mean $\pm$ std) for Pendulum swing-up and Humanoid standup tasks}
  \label{tab:benchmark}
  \begin{tabular}{@{}llccc@{}}
    \toprule
    Task & Algorithm & Total Cost & Plan (ms) & TTG/TTS (s) \\
    \midrule
    \multirow{4}{*}{Pendulum}
      & MPPI & $10.20 \pm 0.34$ & \textbf{6.2} & 3.58$^\dagger$ \\
      & DIAL & $\mathit{8.71 \pm 0.14}$ & 67.7 & 3.23 \\
      & MBD & $9.09 \pm 0.21$ & 70.8 & 3.48$^\dagger$ \\
      & \textbf{MToD} & $\mathbf{8.79 \pm 0.10}$ & 6.4
        & \textbf{3.19}\\
    \midrule
    \multirow{4}{*}{Humanoid}
      & MPPI & $7.43 \pm 1.27$ & \textbf{20.5} & 1.96 \\
      & DIAL & $5.01 \pm 0.69$ & 165.2 & 1.11 \\
      & MBD & $24.72 \pm 18.57$ & 454.1
        & 4.20$^\ddagger$ \\
      & \textbf{MToD} & $\mathbf{4.51 \pm 0.30}$ & 28.5
        & \textbf{0.70} \\
    \bottomrule
  \end{tabular}

  {\footnotesize $^\dagger$Excludes 1/10 seed that did not
    reach the goal within the simulation.\\
    $^\ddagger$Excludes 6/10 seeds that did not stand within
    the simulation.}
\end{table}

Table~\ref{tab:benchmark} compares total cost and planning
time. While MToD and DIAL achieve comparable costs, MToD runs at
${\approx}156$\,Hz---nearly the same rate as vanilla MPPI
($162$\,Hz)---because its online phase is simply warm-started
single-iteration MPPI. In contrast, DIAL and MBD require multiple
optimization iterations per step, reducing their planning rate to
$14$--$15$\,Hz, over $10\times$ slower. This demonstrates a key
advantage of the MToD pipeline: expensive multi-modal exploration is
performed offline, while online execution retains the speed of
a lightweight controller. MToD also exhibits the smallest cost
variance across seeds ($\sigma{=}0.10$), suggesting that the swing
classifier provides robust warm-starts regardless of the specific
random initialization.

Additionally, MToD uses significantly less control energy
($1.54$ vs.\ $2.04$ for DIAL), indicating more efficient
trajectories. MToD also achieves the fastest time-to-goal
($3.19$\,s, computed over seeds that reached the upright
position), outperforming DIAL ($3.23$\,s)
and MPPI ($3.58$\,s). The warm-started reference trajectory
guides the controller toward the goal more directly, reducing
wasted exploratory motion. DIAL achieves the smallest final
angle error ($2.77^\circ$) owing to its iterative refinement at
each step, while MToD's error ($4.62^\circ$) is comparable to
MPPI ($4.77^\circ$), reflecting the single-iteration online
controller. These results highlight a favorable trade-off: MToD
provides DIAL-level cost optimization at MPPI-level planning
speed, with the added benefit of multi-modal trajectory discovery.

\begin{figure}[t]
  \centering
  \includegraphics[width=\columnwidth]{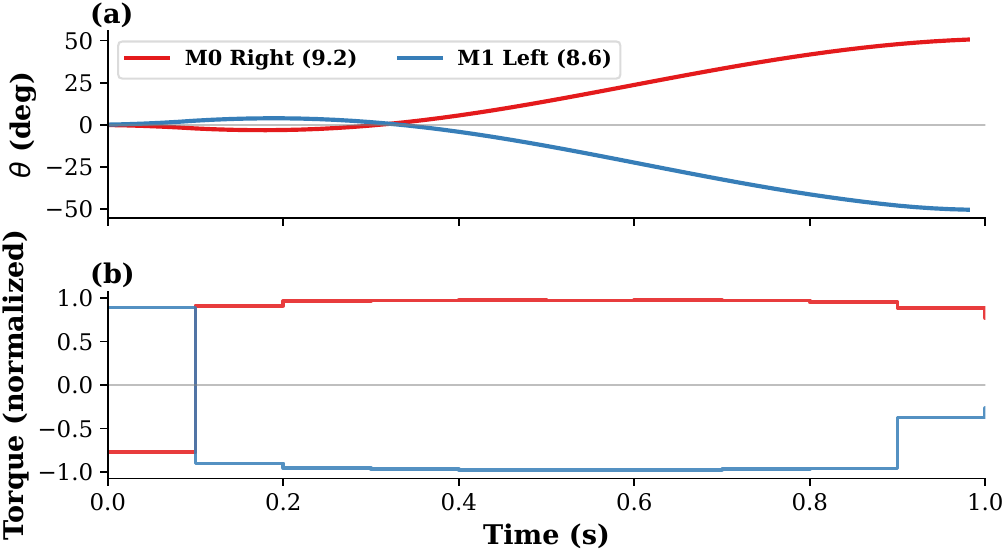}
  \caption{Open-loop mode behavior ($2$~modes, swing classification).
    \emph{(a)}~Angle~$\theta$ trajectories showing right swing
    ($\theta{>}0$) and left swing ($\theta{<}0$). Online cost in
    parentheses.
    \emph{(b)}~Control torque profiles (zero-order hold from knots).
    The two modes exhibit mirror-symmetric torque patterns.}
  \label{fig:pendulum_mode_behavior}
\end{figure}

\noindent\textbf{Physical Mode Behavior.}
Fig.~\ref{fig:pendulum_mode_behavior} visualizes the two open-loop
reference trajectories extracted by the swing classifier for a
representative seed. Panel~(a) shows the angle~$\theta$ over the
$1.0$\,s planning horizon. The two modes cleanly separate into
right swing ($\theta > 0$) and left swing ($\theta < 0$) families,
confirming that the final-angle sign correctly identifies the
committed swing direction despite the energy-pumping transient in
the first ${\approx}0.3$\,s.

Panel~(b) shows the corresponding control torque profiles
(zero-order hold from $N_k{=}11$ knots). The right-swing mode applies
sustained positive torque after an initial negative kick, while
the left-swing mode
mirrors this pattern with sustained negative torque. These distinct
torque signatures confirm that the swing classifier extracts the two
physically meaningful swing-up strategies.

Fig.~\ref{fig:teaser}a shows the corresponding closed-loop
executions rendered at six time instants, with
orange arrows indicating the tip velocity.  Starting from
the downward equilibrium, the right-swing mode drives the pendulum
rightward while the left-swing mode drives it leftward; both reach
the inverted position
by $t{\approx}3.0$\,s and stabilize there.  The vertical dashed line
marks the shared pivot axis, highlighting the mirror symmetry between
the two swing-up strategies---consistent with the $\theta(t)$ and
torque profiles in Fig.~\ref{fig:pendulum_mode_behavior}.

%% file: sections/exp_humanoid.tex
\subsection{Humanoid Standup}
\label{sec:exp:humanoid}

\noindent\textbf{Task.}
We evaluate on a 29-DOF humanoid standup task, where the controller
must drive a humanoid from a supine position on the ground to a stable
standing posture. The high dimensionality ($n_u{=}29$) and rich
contact dynamics make this a significantly more challenging benchmark
than the pendulum. Multiple qualitatively distinct standup strategies
exist---including forward roll, backward push-up, and lateral
roll---making this task a natural test bed for multi-modal trajectory
optimization in high-dimensional systems.

\noindent\textbf{Setup.}
We compare MToD against three online controllers, all running in closed
loop for $8$\,s of simulation:
\begin{itemize}
\item \emph{MPPI} \cite{williams2018itmcp}: vanilla MPPI with 256~samples and
  1~iteration per step, no warm-start.
\item \emph{DIAL-MPC}~\cite{xue2024dialmpc}: diffusion-inspired annealing with
  256~samples and 4~iterations per step.
\item \emph{MBD} \cite{pan2024model}: online Monte Carlo score ascent with 256~samples
  and 10~iterations per step.
\end{itemize}
MToD uses $K{=}128$ DDPM chains ($N{=}256$ diffusion steps each),
score ascent refinement ($50$ iterations, $S{=}256$ samples),
behavioral mode extraction (merge radius $5.0$, budget $M{=}4$), and online MPPI
tracking with the best reference trajectory. The online MPPI uses
the same configuration as the baseline (256~samples, 1~iteration).
We evaluate over $10$ independent random seeds with non-overlapping
seed ranges.

\noindent\textbf{Results.}
Table~\ref{tab:benchmark}
summarizes the benchmark. MToD achieves a total cost of
$4.51 \pm 0.30$, substantially outperforming
MPPI ($7.43 \pm 1.27$), DIAL ($5.01 \pm 0.69$), and MBD online
($24.72 \pm 18.57$). Compared to MPPI, MToD reduces cost by
$39.3\%$; compared to DIAL, by $10.0\%$. MBD online fails to
stand on 6 of 10~seeds, with high variance reflecting the
difficulty of online diffusion-based optimization in this
high-dimensional space. All 10~seeds consistently produce $4$
distinct modes, reflecting the diversity of feasible standup
strategies in the 29-DOF action space.

MToD also achieves the fastest time-to-stand ($0.70$\,s vs.\
$1.11$\,s for DIAL and $1.96$\,s for MPPI), indicating that the
warm-started reference provides an effective initial trajectory
that guides the humanoid toward standing more rapidly.
Notably, MToD exhibits the smallest cost variance
($\sigma{=}0.30$), suggesting that offline mode extraction
provides robust warm-starts, even in this high-dimensional setting.

\noindent\textbf{Physical Mode Behavior.}
Phase~3 projects the $21$-dimensional behavioral features into a
PCA subspace retaining $96\%$ of variance ($14$ components for
seed~$6$) and applies greedy diversity filtering. The resulting
four modes---\emph{Hand-Push}, \emph{Leg-Swing}, \emph{Lateral
Roll}, and \emph{Kip-Up} (Fig.~\ref{fig:teaser}b)---are
well-separated in the decorrelated feature space, confirming that
the behavioral embedding captures task-relevant distinctions that
raw control-space distances (which appear nearly uniform across
chains) do not resolve.

\begin{figure}[t]
  \centering
  \includegraphics[width=\columnwidth]{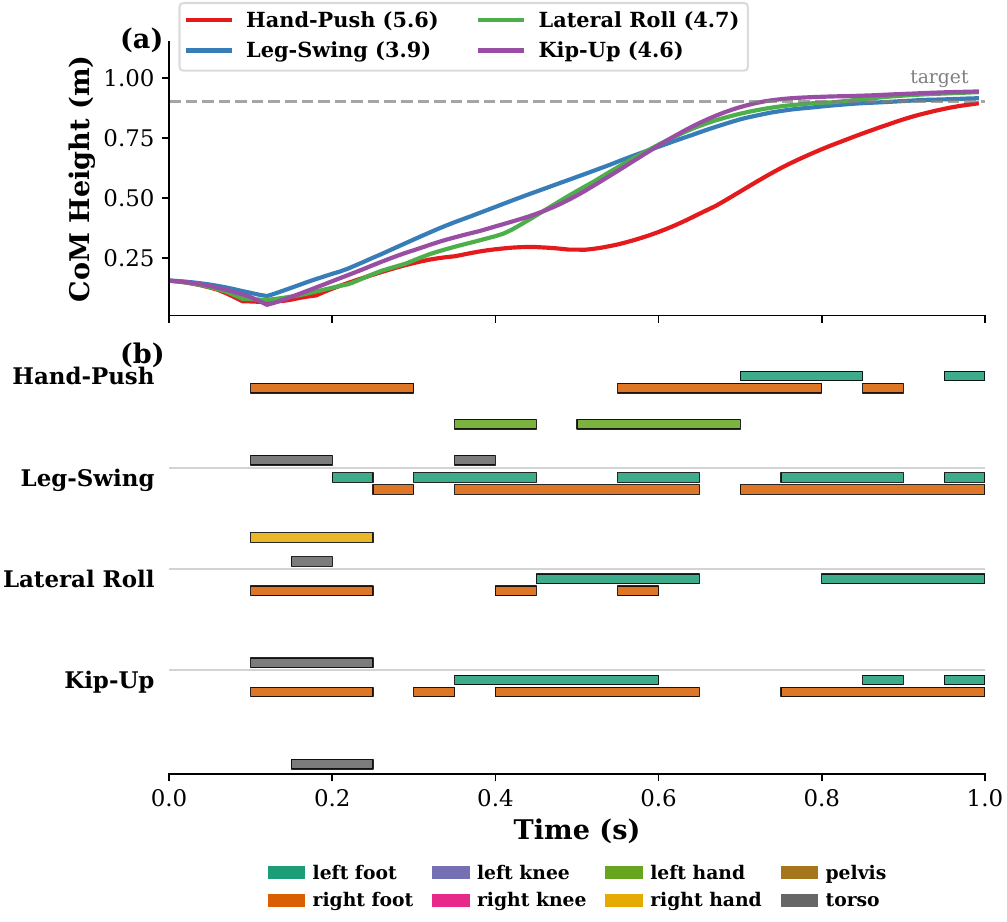}
  \caption{Online closed-loop mode behavior (seed~6, first $1.0$\,s).
    \emph{(a)}~CoM height trajectories for the $4$ extracted modes
    under MPPI tracking (online cost in parentheses). Dashed line
    marks target height.
    \emph{(b)}~Floor contact timeline per mode; colored bars indicate
    body-group contact with the ground. All modes reach the target
    via qualitatively distinct contact sequences.}
  \label{fig:humanoid_mode_behavior}
\end{figure}

Fig.~\ref{fig:humanoid_mode_behavior} visualizes the first $1.0$\,s
of online closed-loop tracking for these four modes. Panel~(a) shows
CoM height trajectories and Panel~(b) presents floor contact
timelines for all eight body-part groups. Under MPPI feedback
control, all four modes approach the target height by $1.0$\,s yet
follow qualitatively distinct standup strategies, each motivating
its name:

\emph{Hand-Push} (red) is the slowest to rise: the CoM remains
below $0.3$\,m until $0.5$\,s while intermittent hand contacts
adjust the body's orientation, after which a rapid push-off drives
the CoM steeply upward, just reaching the target at $1.0$\,s. The
defining hand-support interval in the early phase gives the mode
its name.

\emph{Leg-Swing} (blue) rises fastest among all four modes,
reaching ${\sim}0.85$\,m by $0.5$\,s and stabilizing near the
target thereafter. The early phase features brief alternating foot
taps as the legs swing to accumulate angular momentum, followed by
sustained bilateral foot support from $0.5$\,s onward. This
pendulum-like leg pumping gives the mode its name.

\emph{Lateral Roll} (green) ascends at a moderate pace, with the
CoM rising steadily to ${\sim}0.85$\,m by $1.0$\,s. The contact
timeline shows prolonged early torso contact transitioning to
asymmetric foot placement around $0.6$\,s, forming a lateral
pivot axis around which the body rolls from supine to upright.
The side-rolling support pattern motivates the name
\emph{Lateral Roll}.

\emph{Kip-Up} (purple) exhibits the most explosive initial
dynamics: a rapid CoM surge to ${\sim}0.8$\,m within $0.4$\,s.
The contact chart shows continuous torso sliding in the first
half, transitioning abruptly to bilateral foot support around
$0.5$\,s. This torso-driven explosive standup resembles the
gymnastic kip-up maneuver.

Comparing the four modes, \emph{Leg-Swing} achieves the lowest
online tracking cost despite minimal early contact, suggesting
that its momentum-driven trajectory is easiest for the downstream
MPPI controller to follow. Conversely, \emph{Hand-Push} involves
the most intricate contact sequencing---hand placement timing and
force coordination---resulting in the highest online cost despite
having the lowest offline planning cost. This discrepancy
underscores the value of retaining multiple modes: the strategy
that appears best in open-loop planning is not necessarily the one
that performs best under online feedback control.

%% file: sections/appendix_box_score.tex
\makeatletter
\@ifundefined{proposition}{\newtheorem{proposition}{\bf Proposition}}{}
\makeatother

Throughout, $\mathcal{B}\coloneqq\prod_{k=1}^{N_k}\prod_{d=1}^{n_u}
[u_{\min,d},u_{\max,d}]\subset\mathbb{R}^{D}$, $D\coloneqq N_k n_u$, is
the actuator box, $\mathbb{1}_{\mathcal{B}}$ its indicator,
$\rho(\bU)\coloneqq\exp(-J(\bU)/\lambda)$ the Boltzmann factor, and
$q(\bU^{(i)}\mid\bU^{(0)})$ the VP kernel of Eq.~\eqref{eq:forward}.

\begin{proposition}[Score of the box-constrained target]
\label{prop:box_score}
Let $p_0(\bU^{(0)})=\rho(\bU^{(0)})\,\mathbb{1}_{\mathcal{B}}(\bU^{(0)})/Z_0$
and $p_i(\bU^{(i)})=\int q(\bU^{(i)}\mid\bU^{(0)})\,p_0(\bU^{(0)})\,
\mathrm{d}\bU^{(0)}$. Define the Gaussian
$\varphi_i(\bU^{(0)};\bU^{(i)})\coloneqq\mathcal{N}\bigl(\bU^{(0)};
\bU^{(i)}/\sqrt{\bar{\alpha}_i},\,\tfrac{1-\bar{\alpha}_i}{\bar{\alpha}_i}\bI\bigr)$
and its restriction to the box
$\varphi_i^{\mathrm{box}}\coloneqq\varphi_i\,\mathbb{1}_{\mathcal{B}}/Z_{\mathcal{B}}$,
$Z_{\mathcal{B}}(\bU^{(i)})\coloneqq\int_{\mathcal{B}}\varphi_i$.
Then, for every $\bU^{(i)}\in\mathbb{R}^{D}$,
\begin{align}
  \nabla_{\bU^{(i)}}\log p_i(\bU^{(i)})
  &= \frac{-\bU^{(i)}+\sqrt{\bar{\alpha}_i}\,
     \mathbb{E}\bigl[\bU^{(0)}\mid\bU^{(i)}\bigr]}{1-\bar{\alpha}_i},
  \label{eq:prop_score}\\
  \mathbb{E}\bigl[\bU^{(0)}\mid\bU^{(i)}\bigr]
  &= \frac{\mathbb{E}_{\varphi_i^{\mathrm{box}}}\bigl[\bU^{(0)}\rho(\bU^{(0)})\bigr]}
          {\mathbb{E}_{\varphi_i^{\mathrm{box}}}\bigl[\rho(\bU^{(0)})\bigr]}.
  \label{eq:prop_postmean}
\end{align}
In particular, for i.i.d.\ $\bU_j^{(0)}\sim\varphi_i^{\mathrm{box}}$,
$j=1,\dots,S$, the self-normalized estimator
$\sum_j w_j\bU_j^{(0)}$ with $w_j\propto\rho(\bU_j^{(0)})$ is consistent
for Eq.~\eqref{eq:prop_postmean}, and $Z_{\mathcal{B}}$ does not enter
the weights.
\end{proposition}

\noindent\textit{Proof.}
Write $q_i\coloneqq q(\bU^{(i)}\mid\bU^{(0)})$ and $p_0\coloneqq
p_0(\bU^{(0)})$ under integrals over $\bU^{(0)}$. Since $q_i$ is smooth
in $\bU^{(i)}$ and $p_0$ is integrable, differentiation passes under the
integral, and with $\nabla_{\bU^{(i)}}q_i=q_i\nabla_{\bU^{(i)}}\log q_i$,
\begin{align}
  \nabla_{\bU^{(i)}}\log p_i(\bU^{(i)})
  &= \frac{\int \nabla_{\bU^{(i)}} q_i\;p_0\,\mathrm{d}\bU^{(0)}}
          {\int q_i\;p_0\,\mathrm{d}\bU^{(0)}}
  \nonumber\\
  &= \frac{\int
       \tfrac{-\bU^{(i)}+\sqrt{\bar{\alpha}_i}\,\bU^{(0)}}{1-\bar{\alpha}_i}\,
       q_i\;p_0\,\mathrm{d}\bU^{(0)}}
     {\int q_i\;p_0\,\mathrm{d}\bU^{(0)}}
  \nonumber\\
  &= -\frac{\bU^{(i)}}{1-\bar{\alpha}_i}
     + \frac{\sqrt{\bar{\alpha}_i}}{1-\bar{\alpha}_i}\,
     \frac{\int \bU^{(0)}\,q_i\;p_0\,\mathrm{d}\bU^{(0)}}
          {\int q_i\;p_0\,\mathrm{d}\bU^{(0)}},
  \label{eq:proof_tweedie}
\end{align}
and the last ratio is the posterior mean under the joint density
$q_i\,p_0$, which gives Eq.~\eqref{eq:prop_score}
(Tweedie's formula~\cite{Efron2011Tweedie}); no property of the support
of $p_0$ was used. Completing the square in $\bU^{(0)}$,
$q_i=\bar{\alpha}_i^{-D/2}\varphi_i(\bU^{(0)};\bU^{(i)})$. Substituting
this and $p_0=\rho\,\mathbb{1}_{\mathcal{B}}/Z_0$ into the ratio of
Eq.~\eqref{eq:proof_tweedie}, the constants cancel and
\begin{align}
  \mathbb{E}\bigl[\bU^{(0)}\mid\bU^{(i)}\bigr]
  &= \frac{\int_{\mathcal{B}}\bU^{(0)}\rho(\bU^{(0)})\,\varphi_i\,\mathrm{d}\bU^{(0)}}
          {\int_{\mathcal{B}}\rho(\bU^{(0)})\,\varphi_i\,\mathrm{d}\bU^{(0)}}
  \nonumber\\
  &= \frac{Z_{\mathcal{B}}\,
       \mathbb{E}_{\varphi_i^{\mathrm{box}}}\bigl[\bU^{(0)}\rho(\bU^{(0)})\bigr]}
     {Z_{\mathcal{B}}\,
       \mathbb{E}_{\varphi_i^{\mathrm{box}}}\bigl[\rho(\bU^{(0)})\bigr]},
  \label{eq:proof_is}
\end{align}
which is Eq.~\eqref{eq:prop_postmean}; the ratio of sample means of the
numerator and denominator is the stated self-normalized estimator, and
its consistency follows from the law of large numbers applied to each
mean. \hfill$\square$

\smallskip
\noindent\textit{Remarks.}
Model-based diffusion~\cite{pan2024model} also computes the score
analytically through the posterior mean, but draws candidates from the
unrestricted Gaussian $\varphi_i$ and enforces feasibility through the
weight, so that every candidate outside $\mathcal{B}$ is rolled out and
then discarded; ours differs in that the actuator box is imposed on the
proposal itself, which leaves the importance weights exactly
proportional to the target-to-proposal density ratio, and hence
introduces no bias, while ensuring that every rollout contributes to the
estimate. The fraction of unrestricted candidates that survive is
$Z_{\mathcal{B}}$, a product over control coordinates that decays with
the proposal bandwidth and exponentially in $D$; at the large bandwidths
of the early reverse steps, where $\varphi_i$ extends far beyond the
actuator range and the sampler must explore the feasible set globally,
this fraction becomes prohibitively small, so truncation is what renders
the high-noise stage well posed rather than a mere efficiency
refinement. Truncation is also to be distinguished from projecting
Gaussian draws onto $\mathcal{B}$ before rollout, which concentrates
proposal mass on the boundary and biases the estimate toward the
actuator limits.